\documentclass[10pt,twocolumn,letterpaper]{article}

\usepackage{cvpr}
\usepackage{times}
\usepackage{epsfig}
\usepackage{graphicx}
\usepackage{amsmath}
\usepackage{amssymb}
\usepackage{mathrsfs}
\usepackage{url}
\usepackage{enumitem}
\usepackage{algorithm}
\usepackage{algpseudocode}
\usepackage{dsfont}
\usepackage{multirow}
\usepackage{booktabs}
\usepackage[pagebackref=false,breaklinks=true,letterpaper=true,colorlinks,urlcolor=magenta,citecolor=blue,linkcolor=blue,bookmarks=false]{hyperref}

\usepackage[breaklinks=true,bookmarks=false]{hyperref}

\cvprfinalcopy 

\def\cvprPaperID{****} 

\begin{document}

\title{Unsupervised Deep Tracking}

\author{Ning~Wang$^{1}$~~~~Yibing~Song$^{2}$\thanks{Y. Song and W. Liu are the corresponding authors. This work is done when N. Wang is an intern in Tencent AI Lab. The source code and results are available at \url{https://github.com/594422814/UDT}.}~~~~Chao~Ma$^{3}$~~~~Wengang~Zhou$^{1}$~~~~Wei~Liu$^{2*}$~~~~Houqiang~Li$^{1}$ \\
	$^{1}$ CAS Key Laboratory of GIPAS, University of Science and Technology of China \\ $^{2}$ Tencent AI Lab \\ $^{3}$ MoE Key Lab of Artificial Intelligence, AI Institute, Shanghai Jiao Tong University\\
	{\tt\small wn6149@mail.ustc.edu.cn, dynamicstevenson@gmail.com, chaoma@sjtu.edu.cn}\\ {\tt\small zhwg@ustc.edu.cn, wl2223@columbia.edu, lihq@ustc.edu.cn}
}


\maketitle
\thispagestyle{empty}
\pagestyle{empty}

\begin{abstract}
	We propose an unsupervised visual tracking method in this paper.
	Different from existing approaches using extensive annotated data for supervised learning, our CNN model is trained on large-scale unlabeled videos in an unsupervised manner.
	Our motivation is that a robust tracker should be effective in both the forward and backward predictions (\ie, the tracker can forward localize the target object in successive frames and backtrace to its initial position in the first frame). 
	We build our framework on a Siamese correlation filter network, which is trained using unlabeled raw videos.
	%
	Meanwhile, we propose a multiple-frame validation method and a cost-sensitive loss to facilitate unsupervised learning.
	Without bells and whistles, the proposed unsupervised tracker achieves the baseline accuracy of fully supervised trackers, which require complete and accurate labels during training.
	%
	%
	Furthermore, unsupervised framework exhibits a potential in leveraging unlabeled or weakly labeled data to further improve the tracking accuracy.
	
\end{abstract}

\section{Introduction} \label{sec:intro}



Visual tracking is a fundamental task in computer vision, which aims to localize the target object in the video given a bounding box annotation in the first frame.
The state-of-the-art deep tracking methods \cite{SiamFc,SINT,SASiam,RASNet,SiamRPN,StructSiam,MemTrack,DCFNet,ACT,RTMDNet,luo2019end,luo2014multiple} typically use pretrained CNN models for feature extraction.
These models are trained in a supervised manner, requiring a large quantity of annotated ground-truth labels.
Manual annotations are always expensive and time-consuming, whereas extensive unlabeled videos are readily available on the Internet.
It deserves to investigate how to exploit unlabeled video sequences for visual tracking.


\begin{figure}
	\centering
	\includegraphics[width=8.7cm]{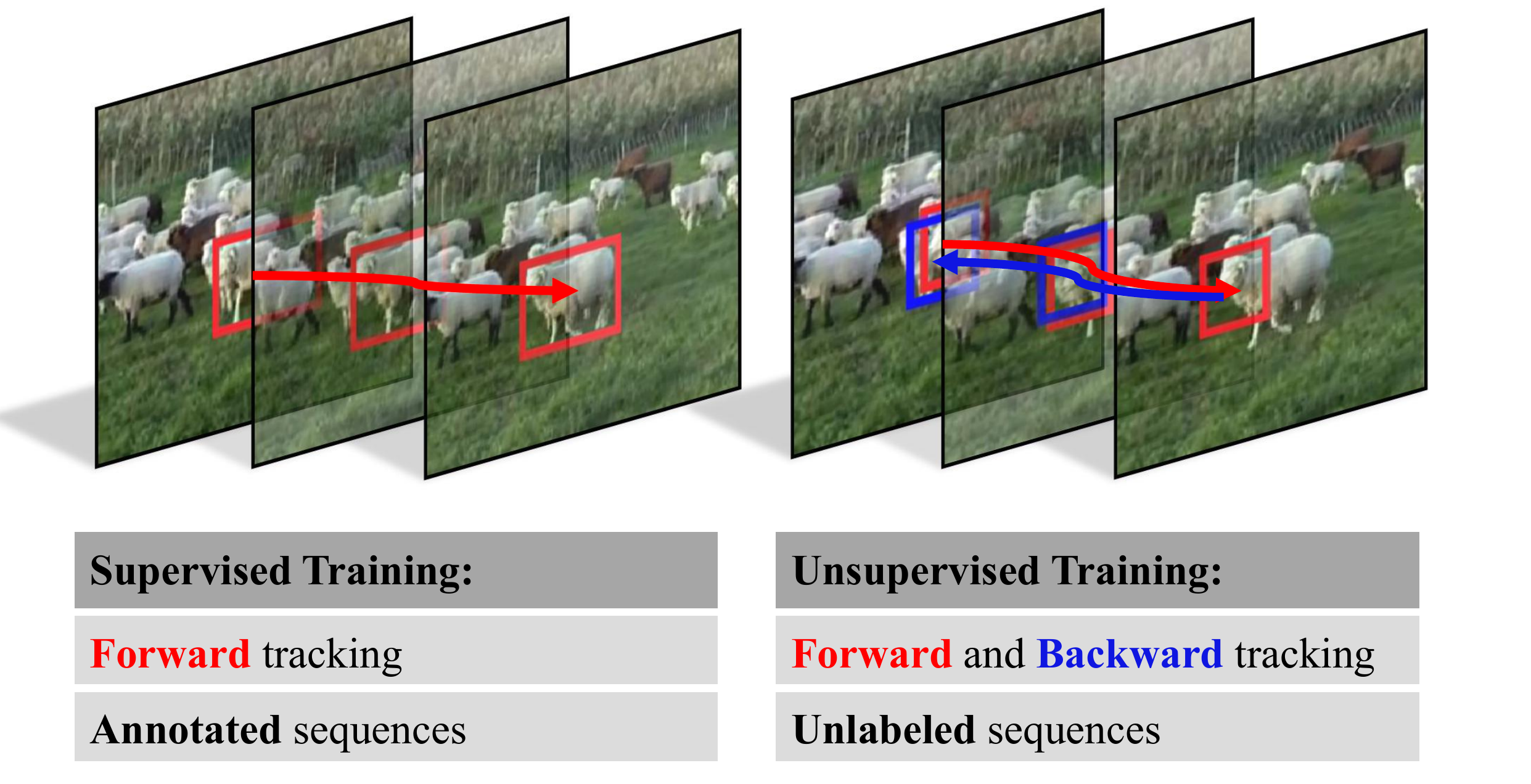}
	\caption{The comparison between supervised and unsupervised learning. Visual tracking methods via supervised learning require ground-truth labels for every frame of the training videos. By utilizing the forward tracking and backward verification, we train the unsupervised tracker without heavyweight annotations.}
	\label{fig:1} \vspace{-0.0in}
\end{figure}

In this paper, we propose to learn a visual tracking model from scratch via unsupervised learning.
Our intuition resides on the observation that visual tracking can be performed in both the forward and backward manners.
Initially, given the target object annotated on the first frame, we can track the target object forward in the subsequent frames.
When tracking backward, we use the predicted location in the last frame as the initial target annotation and track it backward towards the first frame.
The estimated target location in the first frame via backward tracking is expected to be identical with the initial annotation.
After measuring the difference between the forward and backward target trajectories, our network is trained in an unsupervised manner\footnote{In this paper, we do not distinguish between the term \emph{unsupervised} and \emph{self-supervised}, as both refer to learning without ground-truth annotations.} by considering the trajectory consistency as shown in Fig. \ref{fig:1}.
Through exploiting consecutive frames in unlabeled videos, our model learns to locate targets by repeatedly performing forward tracking and backward verification.

The proposed unsupervised learning scheme aims to acquire a generic feature representation, while not being strictly required to track a complete object.
For a video sequence, we randomly initialize a bounding box in the first frame, which may not cover an entire object.
Then, the proposed model learns to track the bounding box region in the following sequences.
This tracking strategy shares similarity with the part-based \cite{Liu_StructuralCF} or edge-based \cite{IBCCF} tracking methods that focus on tracking the subregions of the target objects.
As the visual object tracker is not expected to only concentrate on the complete objects, we use the randomly cropped bounding boxes for tracking initialization during training.


We integrate the proposed unsupervised learning into the Siamese based correlation filter framework \cite{DCFNet}.
The proposed network consists of two steps in the training process: forward tracking and backward verification.
We notice that the backward verification is not always effective since the tracker may successfully return to the initial target location from a deflected or false position.
%
%
In addition, challenges such as heavy occlusion in unlabeled videos will further degrade the network representation capability.
To tackle these issues, we propose multiple frames validation and a cost-sensitive loss to benefit the unsupervised training.
The multiple frames validation increases the discrepancy between the forward and backward trajectories to reduce verification failures.
%
Meanwhile, the cost-sensitive loss mitigates the interference from noisy samples during training.

The proposed unsupervised tracker is shown effective on the benchmark datasets.
Extensive experimental results indicate that without bells and whistles, the proposed unsupervised tracker achieves comparable performance with the baseline fully supervised trackers \cite{SiamFc,CFNet,DCFNet}.
When integrated with additional improvements such as the adaptive online model update \cite{SRDCFdecon,ECO}, the proposed tracker exhibits state-of-the-art performance.
It is worth mentioning that the unsupervised framework shows potential in exploiting unlabeled Internet videos to learn good feature representations for tracking scenarios. Given limited or noisy labels, the unsupervised method exhibits comparable results with the corresponding supervised framework.
In addition, we further improve the tracking accuracy by using more unlabeled data.
%
%
Sec.~\ref{ablation} shows a complete analysis of different training configurations.
%
%
%
%
%
%

In summary, the contributions of our work are three-fold:
\begin{itemize}[noitemsep,nolistsep]	
	\item We propose an unsupervised tracking method based on the Siamese correlation filter backbone, which is learned via forward and backward tracking.
	\item We propose a multiple-frame validation method and a cost-sensitive loss to improve the unsupervised learning performance.		
	\item The extensive experiments on the standard benchmarks show the favorable performance of the proposed method and reveal the potential of unsupervised learning in visual tracking.
\end{itemize}

\def\swone{0.99\linewidth}
\begin{figure*}
	\begin{center}
		\begin{tabular}{c}
			\includegraphics[width=\swone]{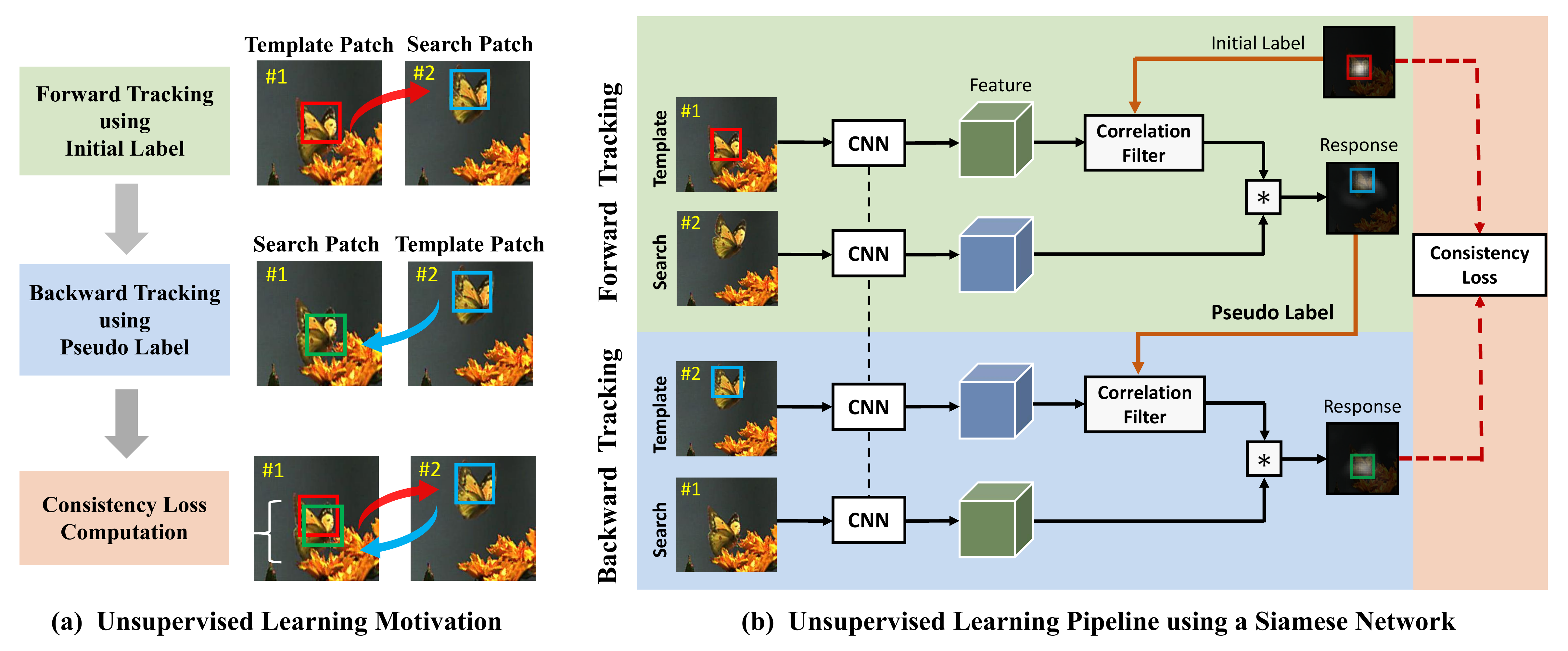}
		\end{tabular}
	\end{center}
	\vspace{-4.5mm}
	\caption{An overview of unsupervised deep tracking. We show our motivation in (a) that we track forward and backward to compute the consistency loss for network training. The detailed training procedure is shown in (b), where unsupervised learning is integrated into a Siamese correlation filter network. Note that during online tracking, we only track forward to predict the target location.}
	\label{fig:2}
\end{figure*}

\section{Related Work}\label{sec:related work}

In this section, we perform a literature review on the deep tracking methods, forward-backward trajectory analysis, and unsupervised representation learning.

{\noindent \bf Deep Visual Tracking.} Existing deep tracking methods either offline learn a specific CNN model for online tracking or simply utilize off-the-shelf deep models (e.g., VGG \cite{VGG,VGGM}) for feature extraction.
%
The Siamese trackers \cite{SiamFc,SINT,CFNet,DCFNet,RASNet,SASiam,SiamRPN,StructSiam,MemTrack} formulate the tracking task as a similarity matching process.
They typically offline learn a tracking network and do not fine-tune the model online.
%
%
On the other hand, some trackers adopt off-the-shelf CNN models as the feature extraction backbone.
They incrementally train binary classification layers \cite{MDNet,VITAL,DATnips} or regression layers \cite{CREST,DSLT} based on the initial frame.
These methods typically achieve high accuracy while consuming a huge computational cost.
%
%
The Discriminative Correlation Filter (DCF) based trackers \cite{MOSSE,KCF,DSST,Liu_StructuralCF,SCT,ningwangTCSVT,Huangjianglei} tackle the tracking task by solving a ridge regression problem using densely sampled candidates, which also benefit from the powerful off-the-shelf deep features (e.g., \cite{HCF,HDT,MCCT,ECO}).
The main distinction is that deep DCF trackers merely utilize off-the-shelf models for feature extraction and do not online train additional layers or fine-tune the CNN models.
%
Different from the above deep trackers using off-the-shelf models or supervised learning, the proposed method trains a network from scratch using unlabeled data in the wild.

{\noindent \bf Forward-Backward Analysis.} The forward-backward trajectory analysis has been widely explored in the literature. The tracking-learning-detection (TLD) \cite{TLD} uses the Kanade-Lucas-Tomasi (KLT) tracker \cite{KLTtracker} to perform forward-backward matching to detect tracking failures. Lee \emph{et al.} \cite{Multihypothesis} proposed to select the reliable base tracker by comparing the geometric similarity, cyclic weight, and appearance consistency between a pair of forward-backward trajectories. However, these methods rely on empirical metrics to identify the target trajectories. In addition, repeatedly performing forward and backward tracking brings in a heavy computational cost for online tracking. Differently, in TrackingNet \cite{2018trackingnet}, forward-backward tracking is used for data annotation and tracker evaluation. In this work, we revisit this scheme to train a deep visual tracker in an unsupervised manner.

{\noindent \bf Unsupervised Representation Learning.} Our framework relates to the unsupervised representation learning. 
In \cite{lee2017unsupervised}, the feature representation is learned by sorting sequences. The multi-layer auto-encoder on large-scale unlabeled data has been explored in \cite{le2013building}. Vondrick \emph{et al.} \cite{vondrick2016anticipating} proposed to anticipate the visual representation of frames in the future. Wang and Gupta \cite{wang2015unsupervised} used the KCF tracker \cite{KCF} to pre-process the raw videos, and then selected a pair of tracked images together with another random patch for learning CNNs using a ranking loss. Our method differs from \cite{wang2015unsupervised} in two aspects. First, we integrate the tracking algorithm into unsupervised training instead of merely utilizing an off-the-shelf tracker as the data pre-processing tool. Second, our unsupervised framework is coupled with a tracking objective function, so the learned feature representation is effective in presenting the generic target objects. In the visual tracking community, unsupervised learning has rarely been touched. To the best of our knowledge, the only related but different approach is the auto-encoder based method \cite{DLT}. However, the encoder-decoder is a general unsupervised framework \cite{olshausen1997sparse}, whereas our unsupervised method is specially designed for tracking tasks.


\section{Proposed Method}\label{sec:proposed approach}

Fig.~\ref{fig:2}(a) shows an example from the \emph{Butterfly} sequence to illustrate forward and backward tracking.
%
In practice, we randomly draw bounding boxes in unlabeled videos to perform forward and backward tracking.
Given a randomly initialized bounding box label, we first track forward to predict its location in the subsequent frames.
Then, we reverse the sequence and take the predicted bounding box in the last frame as the pseudo label to track backward.
%
The predicted bounding box via backward tracking is expected to be identical with the original bounding box in the first frame.
%
We measure the difference between the forward and backward trajectories using the consistency loss for network training.
%
%
An overview of the proposed unsupervised Siamese correlation filter network is shown in Fig.~\ref{fig:2}(b). In the following, we first revisit the correlation filter based tracking framework and then illustrate the details of our unsupervised deep tracking approach.

\subsection{Revisiting Correlation Tracking}\label{revisit CF}

%
The Discriminative Correlation Filters (DCFs) \cite{MOSSE,KCF} regress the input features of a search patch to a Gaussian response map for target localization.
When training a DCF, we select a template patch $ \bf X $ with the ground-truth label $ \bf Y $.
The filter $ \bf W $ can be learned by solving the ridge regression problem as follows:
\begin{equation}\label{Eq1}
\min_{\bf W}{\|{\bf W}\ast{\bf X}-{\bf Y}\|}^{2}_{2}+\lambda{\|{\bf W}\|}^{2}_{2},
\end{equation}
where $ \lambda $ is a regularization parameter and $ \ast $ denotes the circular convolution. Eq.~\ref{Eq1}
can be efficiently calculated in the Fourier domain \cite{MOSSE,DSST,KCF} and the DCF can be computed by
\begin{equation}\label{Eq2}
{\bf W} = \mathscr{F}^{-1}\left( \frac{\mathscr{F}(\bf X)\odot\mathscr{F}^{\star}(Y)}{\mathscr{F}^{\star}(\bf X)\odot\mathscr{F}(X)+\lambda}\right),
\end{equation}
where $\odot$ is the element-wise product, $ \mathscr{F}(\cdot) $ is the Discrete Fourier Transform (DFT), $ \mathscr{F}^{-1}(\cdot) $ is the inverse DFT, and $ \star $ denotes the complex-conjugate operation. In each subsequent frame, given a search patch $ \bf Z $, the corresponding response map $ \bf R $ can be computed in the Fourier domain:
\begin{equation}\label{Eq3}
{\bf R} = {\bf W\ast Z}= {\mathscr F}^{-1} \left( {\mathscr F}^{\star}(\bf W)\odot{\mathscr{F} (\bf Z)} \right).
\end{equation}
%

The above DCF framework starts from learning a target template $\bf W$ using the template patch $\bf X$ and then convolves $\bf W$ with a search patch $\bf Z$ to generate the response. Recently, the Siamese correlation filter network \cite{CFNet,DCFNet} embeds the DCF in a Siamese framework and constructs two shared-weight branches as shown in Fig.~\ref{fig:2}(b). The first one is the template branch which takes a template patch $\bf X$ as input and extracts its features to further generate a target template via DCF. The second one is the search branch which takes a search patch $\bf Z$ as input for feature extraction. The target template is then convolved with the CNN features of the search patch to generate the response map. The advantage of the Siamese DCF network is that both the feature extraction CNN and correlation filter are formulated into an end-to-end framework, so that the learned features are more related to the visual tracking scenarios.


\subsection{Unsupervised Learning Prototype}\label{representation learning}

Given two consecutive frames $P_1$ and $P_2$, we crop the template and search patches from them, respectively. By conducting forward tracking and backward verification, the proposed framework does not require ground-truth labeling for supervised training. The difference between the initial bounding box and the predicted bounding box in $P_1$ will formulate a consistency loss for network learning.
%

{\flushleft \bf Forward Tracking.}
We follow \cite{DCFNet} to build a Siamese correlation filter network to track the initial bounding box region in frame $P_1$.
%
%
After cropping the template patch $ \bf T $ from the first frame $P_1$, the corresponding target template $\bf W_{T}$ can be computed as:
\begin{equation}\label{Eq4}
{\bf W}_{\bf T} = \mathscr{F}^{-1}\left( \frac{\mathscr{F}(\varphi_{\theta}({\bf T}))\odot\mathscr{F}^{\star}({\bf Y}_{\bf T})}{\mathscr{F}^{\star}(\varphi_{\theta}({\bf T}))\odot\mathscr{F}(\varphi_{\theta}({\bf T}))+\lambda} \right),
\end{equation}
where $ \varphi_{\theta}(\cdot) $ denotes the CNN feature extraction operation with trainable network parameters $ \theta $, and $ \bf Y_{T} $ is the label of the template patch $ \bf T $. This label is a Gaussian response centered at the initial bounding box center.
Once we obtain the learned target template ${\bf W}_{\bf T}$, the response map of a search patch $ \bf S $ from frame $P_2$ can be computed by
\begin{equation}\label{Eq5}
{\bf R}_{\bf S} = \mathscr{F}^{-1}(\mathscr{F}^{\star}({\bf W}_{\bf T})\odot\mathscr{F}(\varphi_{\theta}({\bf S}))).
\end{equation}
%
%
If the ground-truth Gaussian label of patch $ \bf S $ is available, the network $ \varphi_{\theta}(\cdot) $ can be trained by computing the $ L_{2} $ distance between $ \bf R_{S} $ and the ground-truth. In the following, we show how to train the network without labels by exploiting backward trajectory verification.

{\flushleft \bf Backward Tracking.}
After generating the response map ${\bf R}_{\bf S}$ for frame $P_2$, we create a pseudo Gaussian label centered at its maximum value, which is denoted by $ {\bf Y}_{\bf S} $.
%
In backward tracking, we switch the role between the search patch and the template patch.
By treating $\bf S$ as the template patch, we generate a target template $\bf W_{S}$ using the pseudo label ${\bf Y}_{\bf S}$.
The target template $\bf W_{S}$ can be learned using Eq.~(\ref{Eq4}) by replacing $\bf T$ with $\bf S$ and replacing ${\bf Y}_{\bf T}$ with ${\bf Y}_{\bf S}$.
Then, we generate the response map ${\bf R}_{\bf T}$ through Eq.~(\ref{Eq5}) by replacing ${\bf W}_{\bf T}$ with $\bf W_{S}$ and replacing $\bf S$ with $\bf T$.
Note that we only use one Siamese correlation filter network to track forward and backward. The network parameters $ \theta $ are fixed during the tracking steps.

{\flushleft \bf Consistency Loss Computation.}
After forward and backward tracking, we obtain the response map ${\bf R}_{\bf T}$. Ideally, ${\bf R}_{\bf T}$ should be a Gaussian label with the peak located at the initial target position. In other words,
$ \bf R_{T} $ should be as similar as the originally given label $ \bf Y_{T} $.
Therefore, the representation network $ \varphi_{\theta}(\cdot) $ can be trained in an unsupervised manner by minimizing the reconstruction error as follows:
\begin{equation}\label{Eq9}
{\cal L}_{\text{un}} = \|{\bf R}_{\bf T}-{\bf Y}_{\bf T}\|^{2}_{2}.
\end{equation}

We perform back-propagation of the computed loss to update the network parameters. During back-propagation, we follow the Siamese correlation filter methods \cite{DCFNet,SACF} to update the network as:
\begin{equation}\label{Eq10}
\begin{aligned}
\frac{\partial {\cal L}_{\text{un}}}{\partial \varphi_{\theta}(\bf T)} &= \mathscr{F}^{-1}\left( \frac{\partial {\cal L}_{\text{un}}}{\partial \left(\mathscr{F}\left(\varphi_{\theta}(\bf T)\right)\right)^{\star}} + \left( \frac{\partial {\cal L}_{\text{un}}}{\partial \left(\mathscr{F}\left(\varphi_{\theta}(\bf T)\right)\right)} \right)^{\star} \right),\\
\frac{\partial {\cal L}_{\text{un}}}{\partial \varphi_{\theta}(\bf S)} &= \mathscr{F}^{-1}\left( \frac{\partial {\cal L}_{\text{un}}}{\partial \left(\mathscr{F}\left(\varphi_{\theta}({\bf S})\right)\right)^{\star}}\right).
\end{aligned}
\end{equation}

\subsection{Unsupervised Learning Improvements}\label{sec:stable training}

The proposed unsupervised learning method constructs the objective function based on the consistency between ${\bf R}_{\bf T}$ and $\bf Y_{T}$. In practice, the tracker may deviate from the target in the forward tracking but still return to the original position during the backward process.
However, the proposed loss function does not penalize this deviation because of the consistent predictions.
Meanwhile, the raw videos may contain uninformative or even corrupted training samples with occlusion that deteriorate the unsupervised learning process.
We propose multiple frames validation and a cost-sensitive loss to tackle these limitations.

\subsubsection{Multiple Frames Validation}
We propose a multiple frames validation approach to alleviate the inaccurate localization issue that is not penalized by Eq.~(\ref{Eq9}). Our intuition is to involve more frames during forward and backward tracking to reduce the verification failures.
The reconstruction error in Eq.~(\ref{Eq9}) tends to be amplified and the computed loss will facilitate the training process.

During unsupervised learning, we involve another frame $P_3$ which is the subsequent frame after $P_2$. We crop a search patch ${\bf S}_{1}$ from $P_2$ and another search patch ${\bf S}_2$ from $P_3$. If the generated response map $ {\bf R}_{{\bf S}_1}$ is different from its corresponding ground-truth response, this error tends to become larger in the next frame $P_3$.
As a result, the consistency is more likely to be broken in the backward tracking, and the generated response map $\bf R_{T}$ is more likely to deviate from $\bf Y_{T}$.
By simply involving more search patches during forward and backward tracking, the proposed consistency loss will be more effective to penalize the inaccurate localizations as shown in Fig.~\ref{fig:3}.
In practice, we use three frames to validate and the improved consistency loss is written as:
\begin{equation}\label{Eq11}
{\cal L}_{\text{un}} = \|\widetilde{\bf R}_{\bf T}-{\bf Y}_{\bf T}\|^{2}_{2},
\end{equation}
where $ \widetilde{\bf R}_{\bf T} $ is the response map generated by an additional frame during the backward tracking step.

\def\swone{0.92\linewidth}
\begin{figure}[t]
	\begin{center}
		\begin{tabular}{c}	
			\includegraphics[width=\swone]{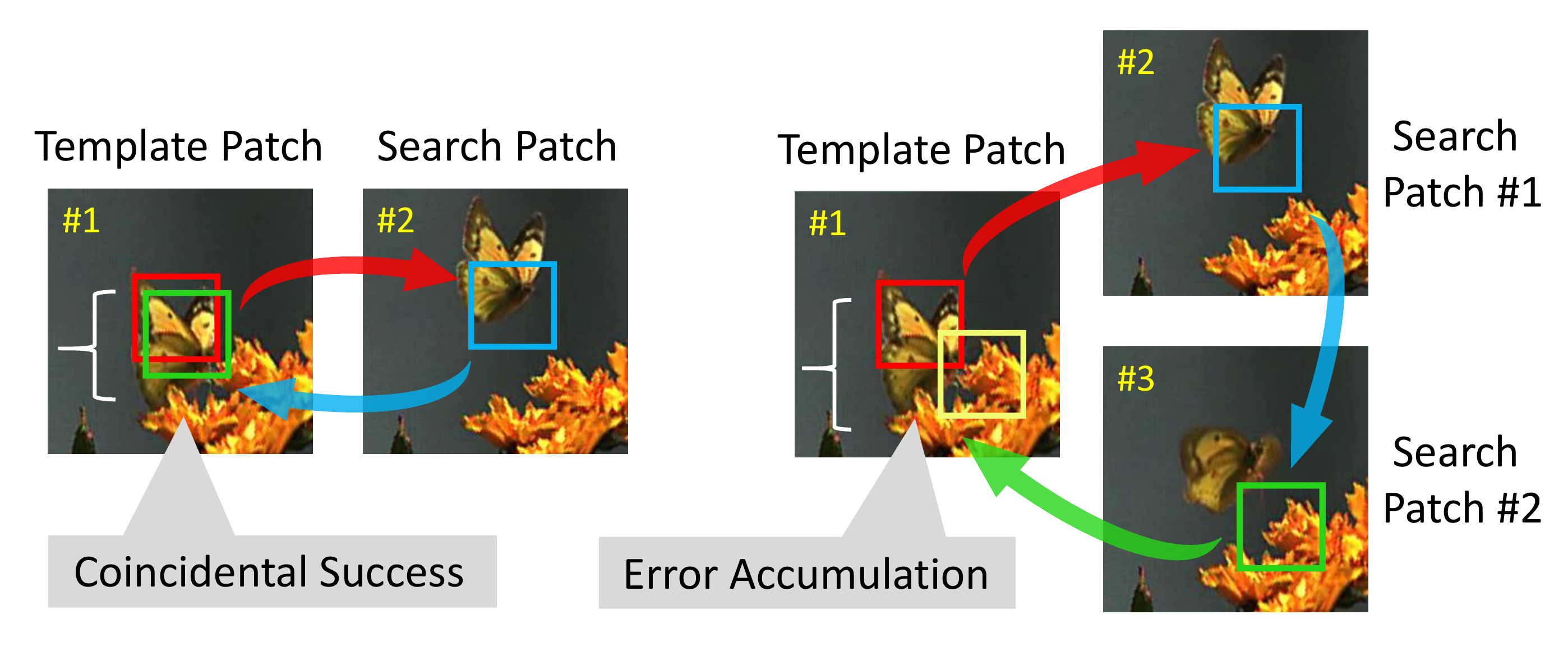}
		\end{tabular}
	\end{center}
	\vspace{-3mm}
	\caption{Single frame validation and multiple frames validation. The inaccurate localization in single frame validation may not be captured as shown on the left. By involving more frames as shown on the right, we can accumulate the localization error to break the prediction consistency during forward and backward tracking.}
	\label{fig:3} \vspace{-0.02in}
\end{figure}

\subsubsection{Cost-sensitive Loss}

We randomly initialize a bounding box region in the first frame $P_1$ for forward tracking.
This bounding box region may contain noisy background context (e.g., occluded targets).
Fig.~\ref{fig:5} shows an overview of these regions.
To alleviate the background interference, we propose a cost-sensitive loss to exclude noisy samples for network training.

During unsupervised learning, we construct multiple training pairs from the training sequences. Each training pair consists of one initial template patch $\bf T$ in frame $P_1$ and two search patches ${\bf S}_{1}$ and ${\bf S}_{2}$ from the subsequent frames $P_2$ and $P_3$, respectively. These training pairs form a training batch to train the Siamese network.
In practice, we find that few training pairs with extremely high losses prevent the network training from convergence.
To reduce the contributions of noisy pairs, we exclude 10\% of the whole training pairs which contain a high loss value. Their losses can be computed using Eq.~(\ref{Eq11}).
To this end, we assign a binary weight ${\bf A}^{i}_{\text{drop}}$ to each training pair and all the weight elements form the weight vector $\bf A_{\text{drop}}$. The 10\% of its elements are 0 and the others are 1.

In addition to the noisy training pairs, the raw videos include lots of uninformative image patches which only contain the background or still targets. For these patches, the objects (e.g., sky, grass, or tree) hardly move.
Intuitively, the target with a large motion contributes more to the network training. 
Therefore, we assign a motion weight vector $\bf A_{\text{motion}}$ to all the training pairs.
Each element ${\bf A}_{\text{motion}}^i$ can be computed by
\begin{equation}\label{Eq12}
{\bf A}_{\text{motion}}^i = \left\|{\bf R}_{{\bf S}_{1}}^i-{\bf Y}_{\bf T}^i\right\|^{2}_{2} +\left\|{\bf R}_{{\bf S}_{2}}^i-{\bf Y}_{{\bf S}_{1}}^i\right\|^{2}_{2},
\end{equation}
where $ {\bf R}_{{\bf S}_{1}}^i $ and $ {\bf R}_{{\bf S}_{2}}^i $ are the response maps in the $i$-th training pair, $ {\bf Y}_{{\bf T}}^i $ and $ {\bf Y}_{{\bf S}_{1}}^i $ are the corresponding initial and pseudo labels, respectively. Eq.~(\ref{Eq12}) calculates the target motion difference from frame $P_1$ to $P_2$ and $P_2$ to $P_3$. The larger value of ${\bf A}^{i}_{\text{motion}}$ indicates that the target undergoes a larger movement in this continuous trajectory. On the other hand, we can interpret that the large value of ${\bf A}^{i}_{\text{motion}}$ represents the hard training pair which the network should pay more attentions to. We normalize the motion weight and the binary weight as follows,
%
\begin{equation}\label{Eq13}
{\bf A}_{\text{norm}}^{i} = \frac{{\bf A}^{i}_{\text{drop}}\cdot{\bf A}^{i}_{\text{motion}}}{\sum_{i=1}^{n}{\bf A}^{i}_{\text{drop}}\cdot{\bf A}^{i}_{\text{motion}}},
\end{equation}
where $ n $ is number of the training pairs in a mini-batch. The final unsupervised loss in a mini-batch is computed as:
\begin{equation}\label{Eq14}
{\cal L}_{\text{un}} = \frac{1}{n}\sum_{i=1}^{n}{\bf A}^{i}_{\text{norm}}\cdot\left\|\widetilde{\bf R}_{\bf T}^{i}-{\bf Y}_{\bf T}^{i}\right\|^{2}_{2}.
\end{equation}

\begin{figure}
	\centering
	\includegraphics[width=7.9cm]{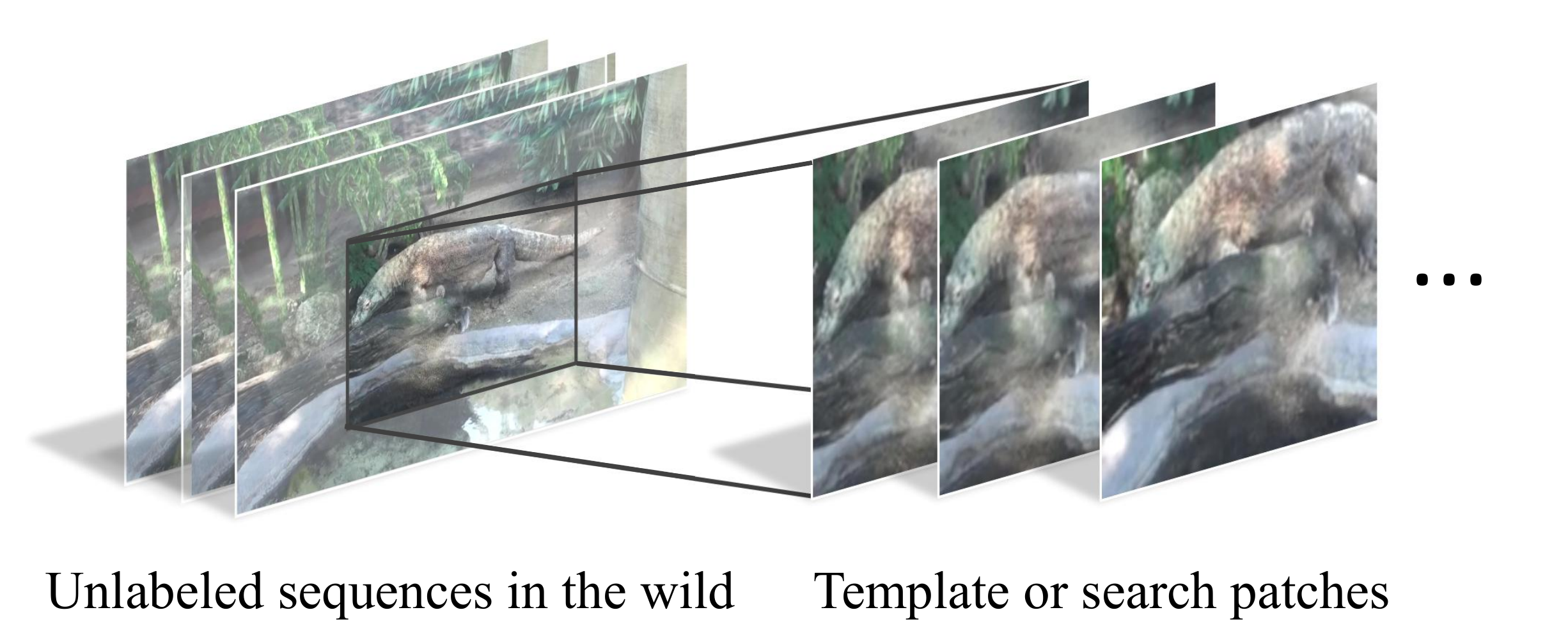}
	\caption{An illustration of training samples generation. The proposed method simply crops and resizes the center regions from unlabeled videos as the training patches.}
	\label{fig:4} \vspace{-0.02in}
\end{figure}

\subsection{Unsupervised Training Details} \label{training details}
{\flushleft \bf Network Structure.} We follow the DCFNet \cite{DCFNet} to use a shallow Siamese network with only two convolutional layers. The filter sizes of these convolutional layers are $3\times3\times3\times32 $ and $ 3\times3\times32\times32 $, respectively. Besides, a local response normalization (LRN) layer is employed at the end of convolutional layers. This lightweight structure enables extremely efficient online tracking.

{\flushleft \bf Training Data.} We choose the widely used ILSVRC 2015 \cite{ILSVRC2015} as our training data to fairly compare with existing supervised trackers. In the data pre-processing step, existing supervised approaches \cite{SiamFc,CFNet,DCFNet} require ground-truth labels for every frame. Meanwhile, they usually discard the frames where the target is occluded, or the target is partially out of view, or the target infrequently appears in tracking scenarios (e.g., snake). This requires a time-consuming human interaction to preprocess the training data. 

\begin{figure}
	\centering
	\includegraphics[width=7.9cm]{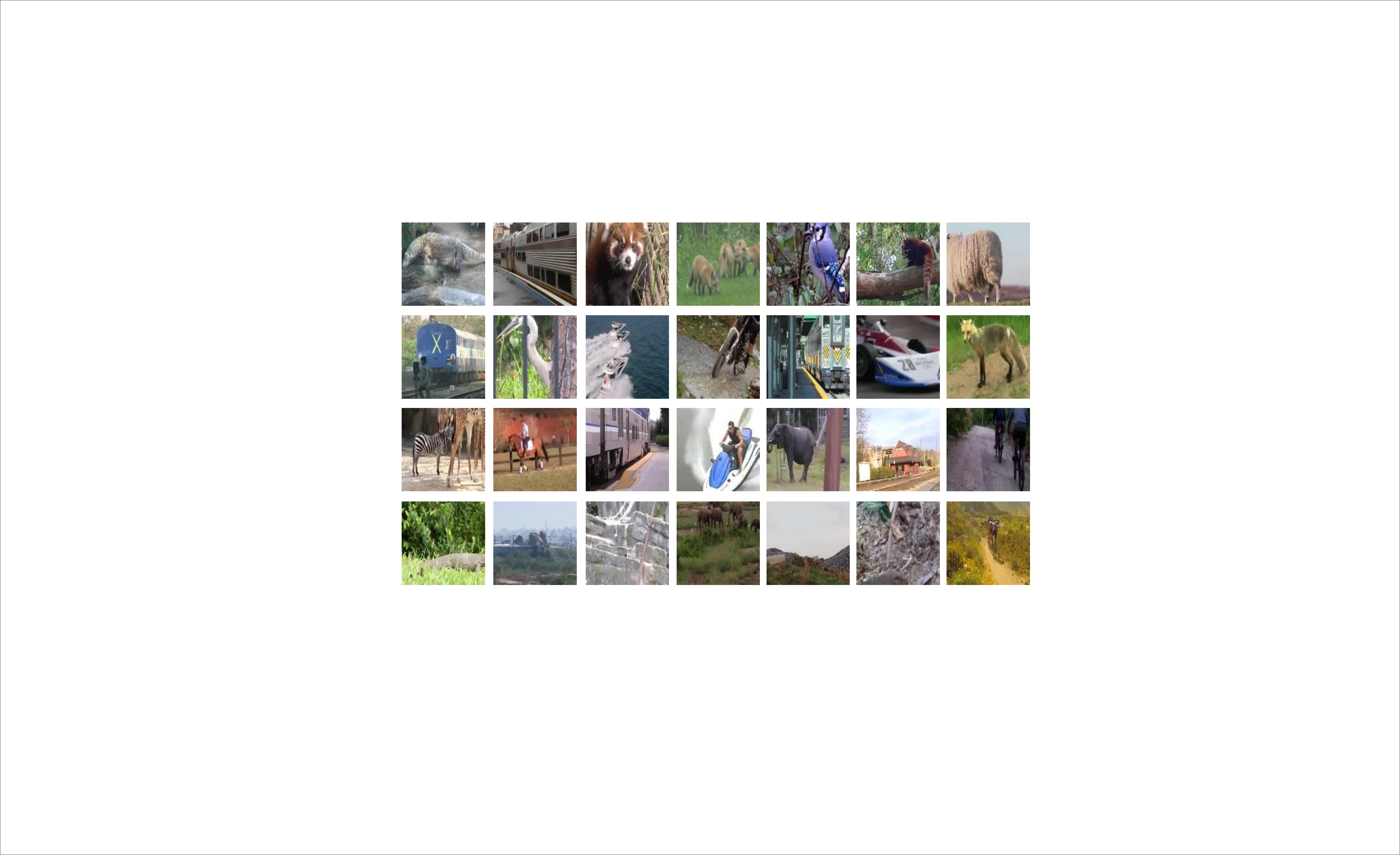}
	\caption{Examples of randomly cropped center patches from ILSVRC 2015 \cite{ILSVRC2015}. Most patches contain valuable contents while some are less meaningful (e.g., the patches on the last row).}
	\label{fig:5} \vspace{-0.03in}
\end{figure}

In contrast, we do not preprocess any data and simply crop the center patch in each frame. The patch size is the half of the whole image and further resized to $125\times125$ as the network input as shown in Fig.~\ref{fig:4}. We randomly choose three cropped patches from the continuous 10 frames in a video. We set one of the three patches as the template and the remaining as search patches. This is based on the assumption that the center located target objects are unlikely to move out of the cropped region in a short period. We track the objects appearing in the center of the cropped regions, while not specifying their categories. Some examples of the cropped regions are exhibited in Fig.~\ref{fig:5}.


\subsection{Online Object Tracking}\label{online tracking}
After offline unsupervised learning, we online track the target object following forward tracking as illustrated in Sec.~\ref{representation learning}. To adapt the object appearance variations, we online update the DCF parameters as follows:
\begin{equation}\label{Eq15}
{\bf W}_{t} = (1-\alpha_{t}){\bf W}_{t-1}+\alpha_{t}{\bf W},
\end{equation}
where $ \alpha_{t}\in[0,1] $ is the linear interpolation coefficient. The target scale is estimated through a patch pyramid with scale factors $ \{ a^{s}|a=1.015, s=\{-1, 0, 1\}\} $ following \cite{SRDCF}. We denote the proposed Unsupervised Deep Tracker as UDT, which merely uses standard incremental model update and scale estimation.
%
Furthermore, we use an advanced model update that adaptively changes $ \alpha_{t} $ as well as a better DCF formulation following \cite{ECO}. The improved tracker is denoted as UDT+.


\section{Experiments}\label{experiment}

In this section, we first analyze the effectiveness of our unsupervised learning framework. Then, we compare with state-of-the-art trackers on the standard benchmarks including OTB-2015 \cite{OTB-2015}, Temple-Color \cite{TempleColor128} and VOT-2016 \cite{VOT2016}.

\subsection{Experimental Details} \label{sec:Setup}
In our experiments, we use the stochastic gradient descent (SGD) with a momentum of 0.9 and a weight decay of 0.005 to train our model. Our unsupervised network is trained for 50 epoches with a learning rate exponentially decays from $ 10^{-2} $ to $ 10^{-5} $ and a mini-batch size of 32. All the experiments are executed on a computer with 4.00GHz Intel Core I7-4790K and NVIDIA GTX 1080Ti GPU. 

On the OTB-2015 \cite{OTB-2015} and TempleColor \cite{TempleColor128} datasets, we use one-pass evaluation (OPE) with distance precision (DP) at 20 pixels and the area-under-curve (AUC) of the overlap success plot. On the VOT2016 \cite{VOT2016}, we measure the performance using the Expected Average Overlap (EAO). 

\subsection{Ablation Study and Analysis}\label{ablation}

\begin{figure}
	\centering
	\includegraphics[width=4.12cm]{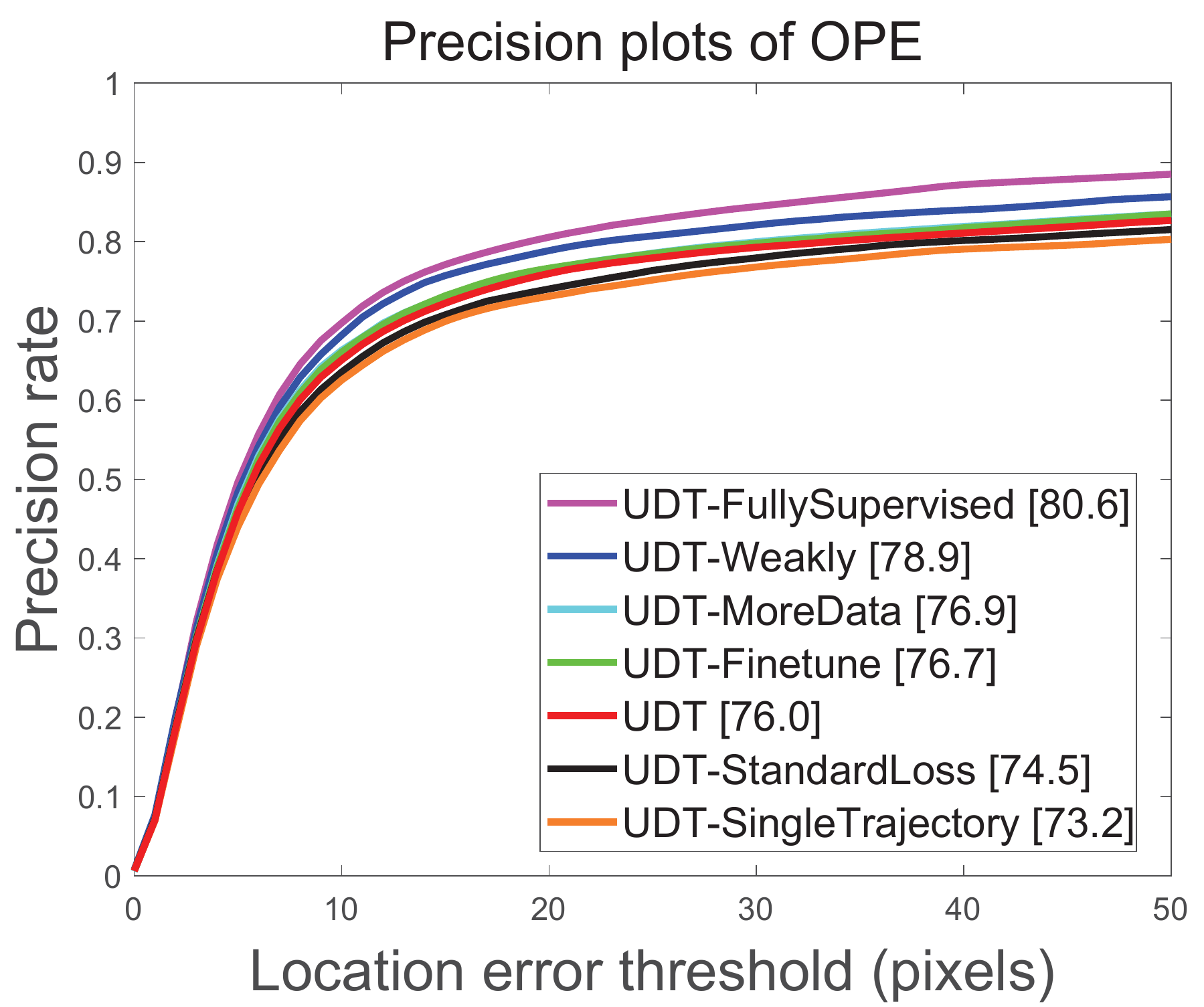}
	\includegraphics[width=4.12cm]{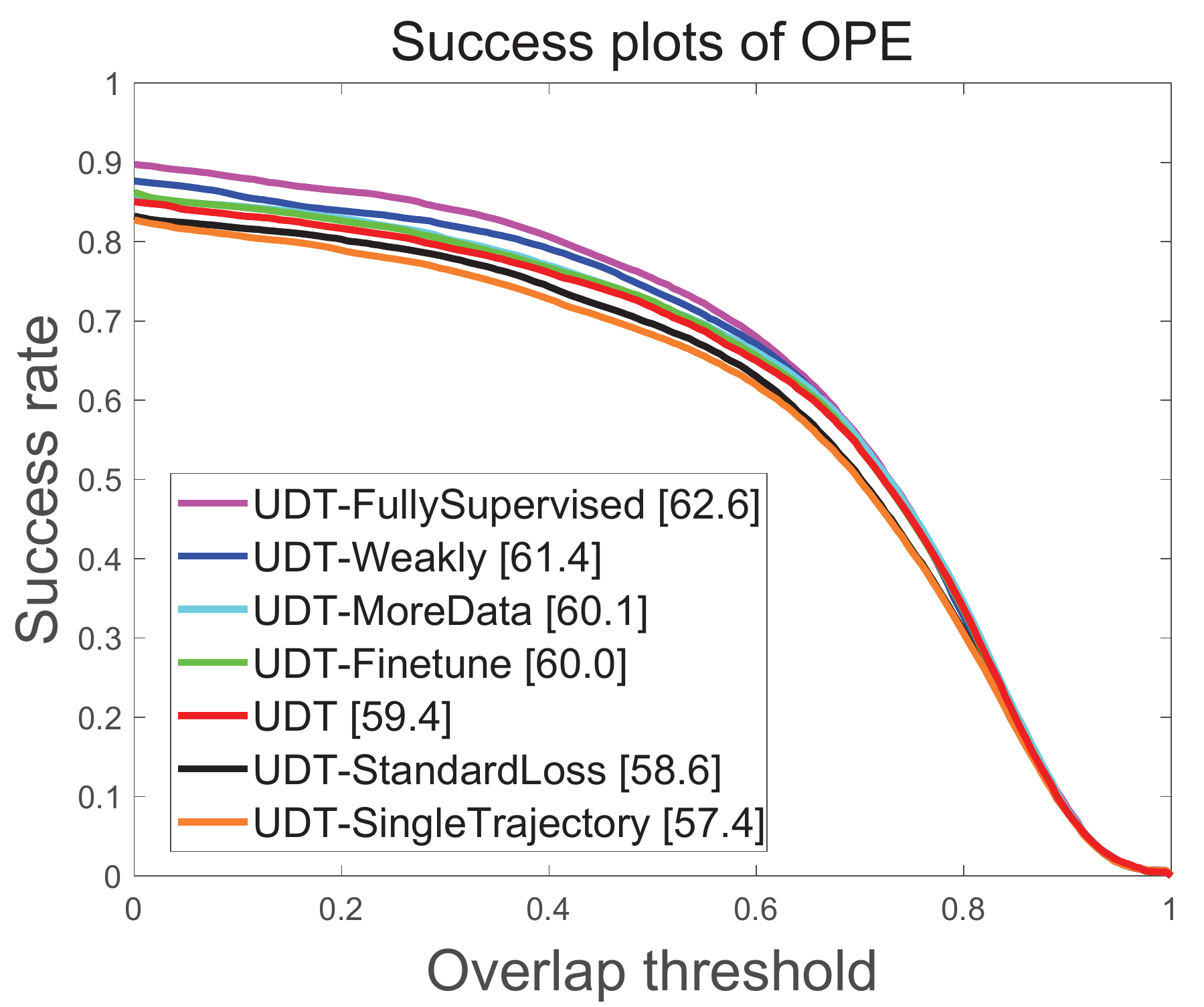}
	\caption{The precision and success plots of our UDT tracker with different configurations on the OTB-2015 dataset \cite{OTB-2015}. In the legend, we show the distance precision at 20 pixels threshold and area-under-curve (AUC) score.} \label{fig:6} \vspace{-0.05in}
\end{figure}

\setlength{\tabcolsep}{2pt}
\begin{table*}
	\scriptsize
	\begin{center}
		\caption{Comparison results with fully-supervised baseline (left) and state-of-the-art (right) trackers on the OTB-2015 benchmark \cite{OTB-2015}. The evaluation metric is AUC score. Our unsupervised UDT tracker performs favorably against baseline methods shown on the left, while our UDT+ tracker achieves comparable results with the recent state-of-the-art supervised trackers shown on the right.} \label{table:table1}
		\vspace{+0.1in}
		\begin{tabular*}{17.3 cm} {@{\extracolsep{\fill}}|l|cccc|cccccccccccc|}
			\hline
			~Trackers &SiamFC &DCFNet  & CFNet  &  {\bf UDT} &DSiam & EAST &HP &SA-Siam & SiamPRN &RASNet &SACF &Siam-tri &RT-MDNet &MemTrack & StructSiam &  {\bf UDT+} \\
			& \cite{SiamFc} &\cite{DCFNet}  & \cite{CFNet}  & &\cite{Dsiam} &\cite{EAST} & \cite{HP} &\cite{SASiam} & \cite{SiamRPN} &\cite{RASNet} &\cite{SACF} &\cite{Siamtriplet}&\cite{RTMDNet} &\cite{MemTrack} &\cite{StructSiam} & \\
			\hline
			\hline
			~AUC score (\%) &58.2  &58.0 &56.8 & 59.4 &60.5 &62.9 &60.1  &65.7  &63.7 &64.2 &63.3 &59.2 &65.0 &62.6 &62.1 &63.2\\
			~Speed (FPS)  &86  &70 &65 &70 &25 &159 &69 &50  &160 &83 &23 &86 &50 &50 &45 &55\\
			\hline
		\end{tabular*}
	\end{center}
	\vspace{-0.07in}
\end{table*}

{\noindent \bf Unsupervised and supervised learning}. We use the same training data \cite{ILSVRC2015} to train our network via fully supervised learning. Fig.~\ref{fig:6} shows the evaluation results where the fully supervised training configuration improves UDT by 3\% under the AUC scores.

{\noindent \bf Stable training.} We analyze the effectiveness of our stable training by using different configurations. Fig.~\ref{fig:6} shows the evaluation results of multiple learned trackers. The UDT-StandardLoss indicates the results from the tracker learned without using hard sample reweighing (i.e., ${\bf A}_{\text{motion}}$ in Eq.~(\ref{Eq12})). The UDT-SingleTrajectory denotes the results from the tracker learned only using the prototype framework in Sec.~\ref{representation learning}. The results show that multiple frames validation and cost-sensitive loss improve the accuracy.


{\noindent \bf Using high-quality training data.}
%
%
We analyze the performance variations by using high-quality training data. In ILSVRC 2015 \cite{ILSVRC2015}, instead of randomly cropping patches, we add offsets ranging from [-20, +20] pixels to the ground-truth bounding boxes for training samples collection. These patches contain more meaningful objects than the randomly cropped ones.
The results in Fig.~\ref{fig:6} show that our tracker learned using weakly labeled samples (i.e., UDT-Weakly) produce comparable results with the supervised configuration.
Note that the predicted target location by existing object detectors or optical flow estimators is normally within 20 pixels offset with respect to the ground-truth. These results indicate that UDT achieves comparable performance with supervised configuration when using less accurate labels produced by existing detection or flow estimation methods.

{\noindent \bf Few-shot domain adaptation.} We collect the first 5 frames from the videos in OTB-2015 \cite{OTB-2015} with only the ground-truth bounding box available in the first frame. Using these limited samples, we fine-tune our network by 100 iterations using the forward-backward pipeline. This training process takes around 6 minutes. The results (i.e., UDT-Finetune) show that the performance is further enhanced. Our offline unsupervised training learns general feature representation, which can be transferred to a specific domain (e.g., OTB) using few-shot adaptation. This domain adaptation is similar to MDNet \cite{MDNet} but our initial parameters are offline learned in an unsupervised manner.

{\noindent \bf Adopting more unlabeled data.} Finally, we utilize more unlabeled videos for network training. These additional raw videos are from the OxUvA benchmark \cite{2018longtermBenchmark} (337 videos in total), which is a subset of Youtube-BB \cite{youtubeBB}. In Fig.~\ref{fig:6}, our UDT-MoreData tracker gains performance improvement (0.9\% DP and 0.7\% AUC), which illustrates unlabeled data can advance the unsupervised training. 
Nevertheless, in the following we remain using the UDT and UDT+ trackers which are only trained on \cite{ILSVRC2015} for fair comparisons.

\subsection{State-of-the-art Comparison}\label{state-of-the-art}
\begin{figure}
	\centering
	\includegraphics[width=4.1cm]{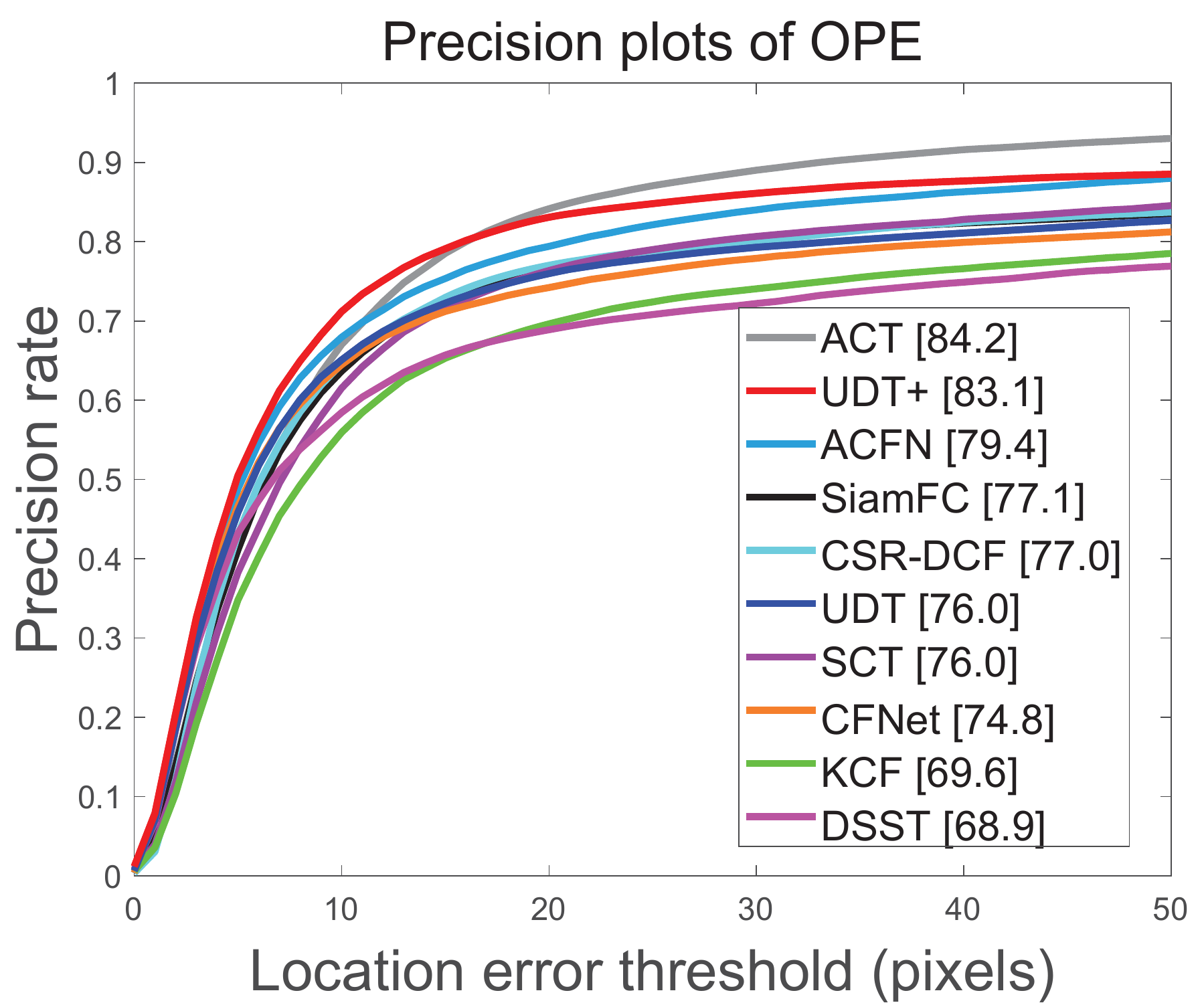}
	\includegraphics[width=4.1cm]{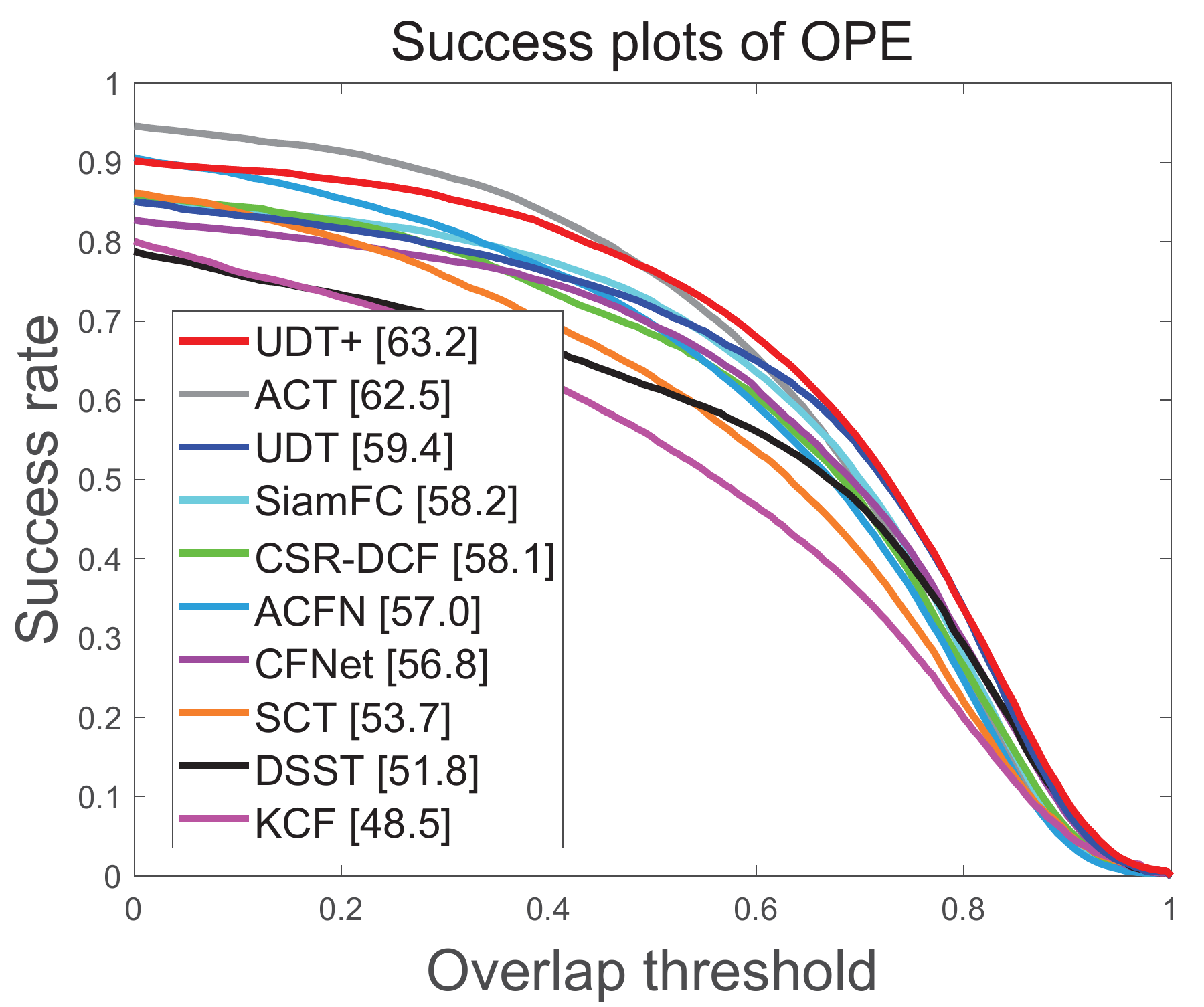}
	\caption{Precision and success plots on the OTB-2015 dataset \cite{OTB-2015} for recent real-time trackers.} \label{fig:7} \vspace{+0.02in}
\end{figure}

{\noindent \bf OTB-2015 Dataset.} 
We evaluate the proposed UDT and UDT+ trackers with state-of-the-art real-time trackers including ACT \cite{ACT}, ACFN \cite{ACFN}, CFNet \cite{CFNet}, SiamFC \cite{SiamFc}, SCT \cite{SCT}, CSR-DCF \cite{CSR-DCF}, DSST \cite{DSST}, and KCF \cite{KCF} using precision and success plots metrics. Fig.~\ref{fig:7} and Table~\ref{table:table1} show that the proposed unsupervised tracker UDT is comparable with the baseline supervised methods (i.e., SiamFC and CFNet). Meanwhile, the proposed UDT tracker exceeds DSST algorithm by a large margin. As DSST is a DCF based tracker with accurate scale estimation, the performance improvement indicates that our unsupervised feature representation is more effective than empirical features.
In Fig.~\ref{fig:7} and Table~\ref{table:table1}, we do not compare with some remarkable non-realtime trackers. For example, MDNet \cite{MDNet} and ECO \cite{ECO} can yield 67.8\% and 69.4\% AUC on the OTB-2015 dataset, but they are far from real-time.

\begin{figure}
	\centering
	\includegraphics[width=4.1cm]{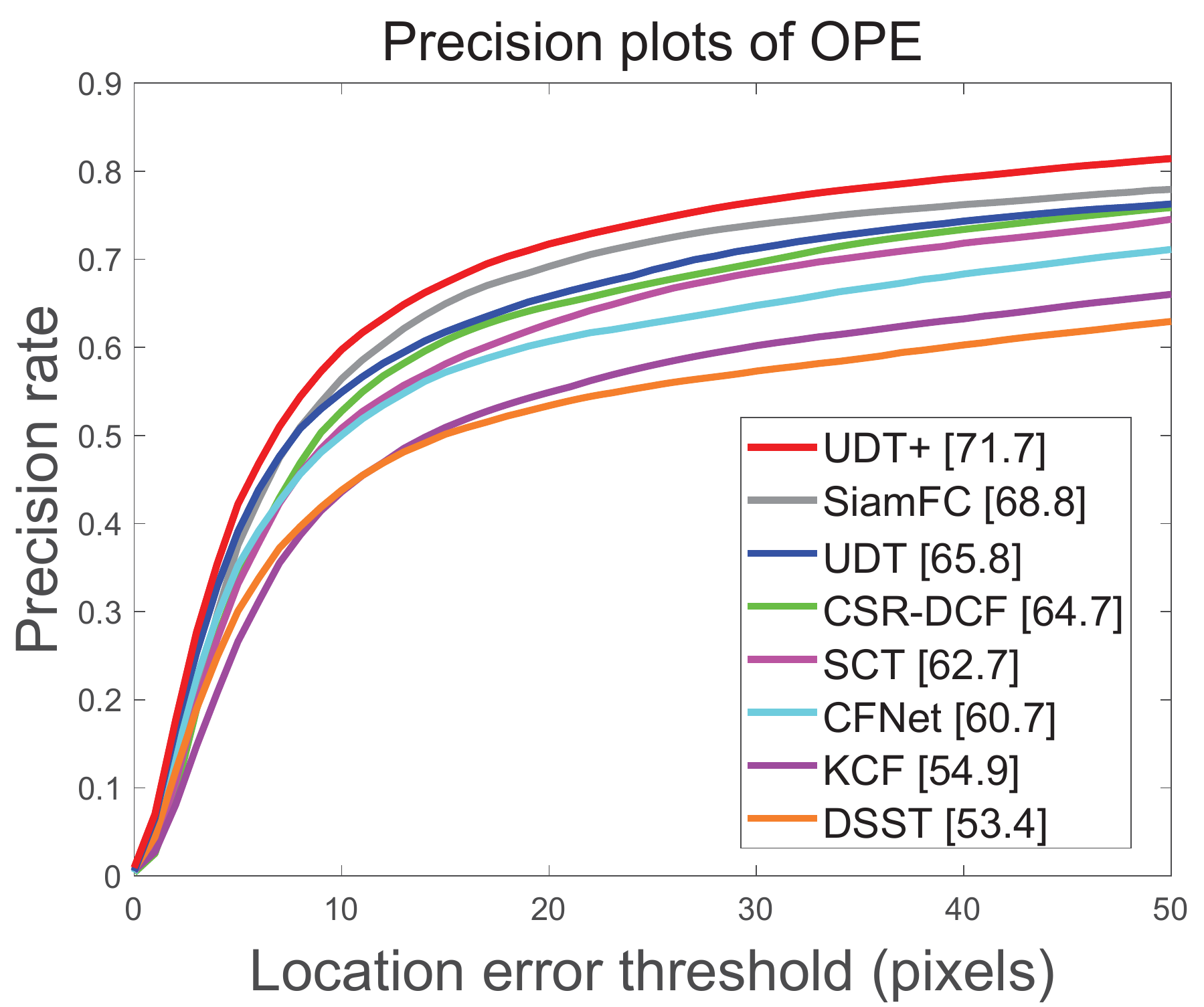}
	\includegraphics[width=4.1cm]{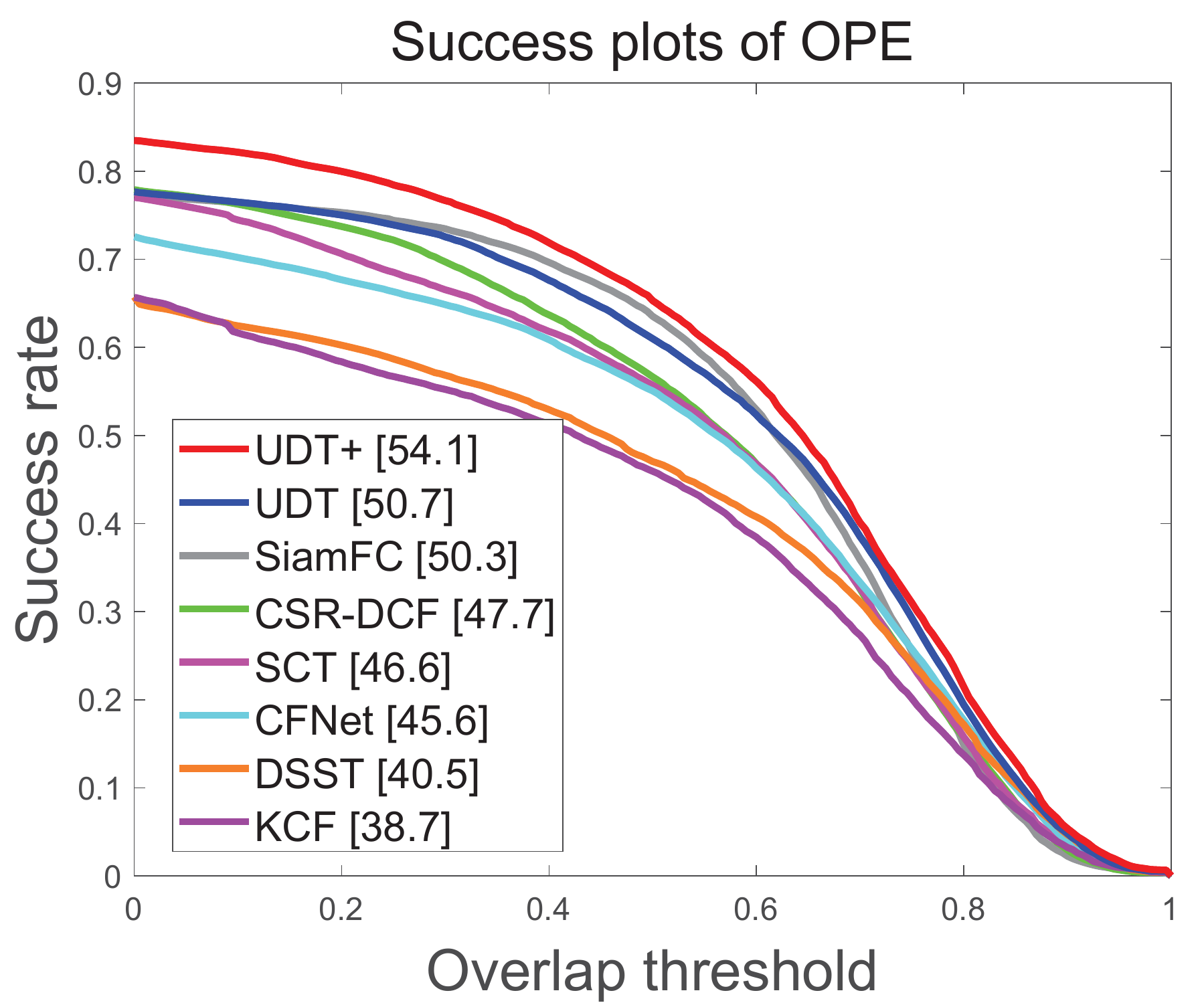}
	\caption{Precision and success plots on the Temple-Color dataset \cite{TempleColor128} for recent real-time trackers.} \label{fig:8} \vspace{-0.05in}
\end{figure}

In Table~\ref{table:table1}, we also compare with more recently proposed supervised trackers. These latest approaches are mainly based on the Siamese network and trained using ILSVRC \cite{ILSVRC2015}. Some trackers (e.g., SA-Siam \cite{SASiam} and RT-MDNet \cite{RTMDNet}) adopt pre-trained CNN models (e.g., AlexNet \cite{Alexnet} and VGG-M \cite{VGGM}) for network initialization. The SiamRPN \cite{SiamRPN} additionally uses more labeled training videos from Youtube-BB dataset \cite{youtubeBB}. Compared with existing methods, the proposed UDT+ tracker does not require data labels or off-the-shelf deep models while still achieving comparable performance and efficiency.


\begin{figure}
	\centering
	\includegraphics[width=8cm]{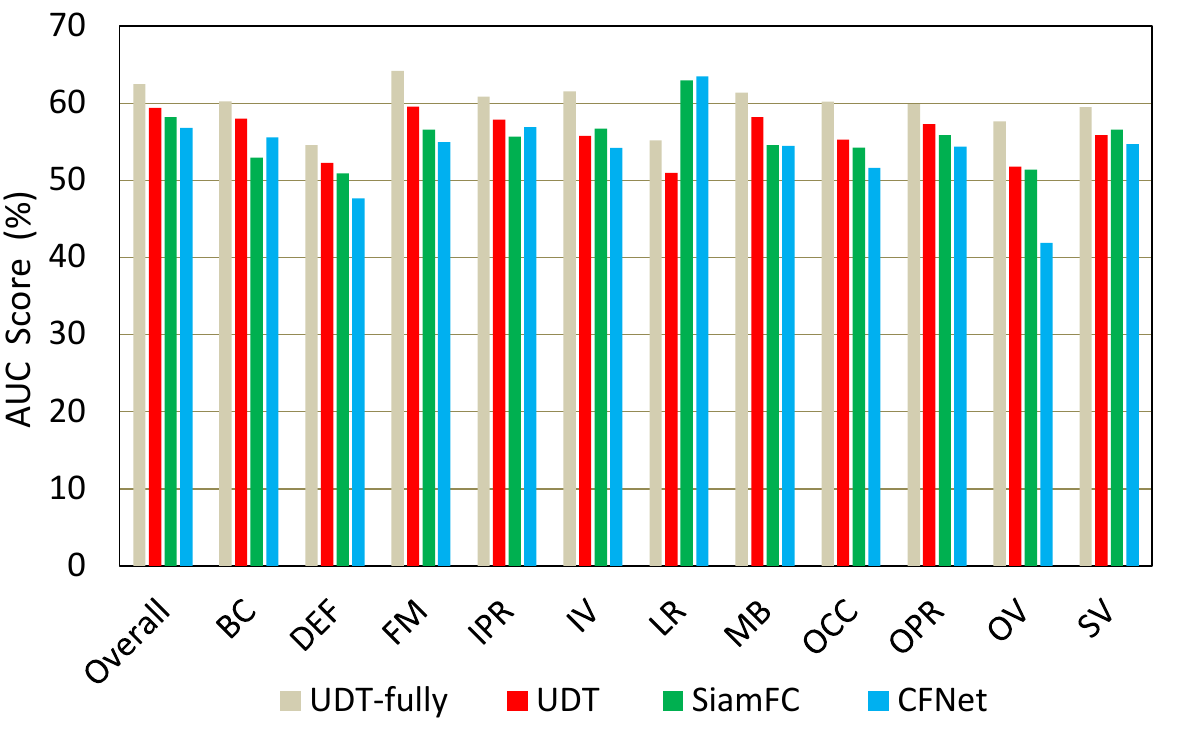}
	\caption{Attribute-based evaluation on the OTB-2015 dataset \cite{OTB-2015}. The 11 attributes are background clutter (BC), deformation (DEF), fast motion (FM), in-plane rotation (IPR), illumination varition (IV), low resolution (LR), motion blur (MB), occlusion (OCC), out-of-plane rotation (OPR), out-of-view (OV), and scale varition (SV), respectively.}
	\label{fig:9} \vspace{+0.05in}
\end{figure}

\setlength{\tabcolsep}{2pt}
\begin{table}
	\footnotesize
	\begin{center}
		\caption{Comparison with state-of-the-art and baseline trackers on the VOT2016 benchmark \cite{VOT2016}. The evaluation metrics include Accuracy, Failures (over 60 sequences), and Expected Average Overlap (EAO). The up arrows indicate that higher values are better for the corresponding metric and vice versa. } \label{table:table2}	
		\vspace{+0.1in}
		\begin{tabular*}{7.5 cm} {@{\extracolsep{\fill}}|l|ccc|c|}
			\hline
			~~Trackers & Accuracy ($ \uparrow $) & Failures ($ \downarrow $) & EAO ($ \uparrow $)~ & FPS ($ \uparrow $)~ \\
			\hline
			\hline
			~~ECO \cite{ECO} &0.54 &- &0.374 &6\\
			~~C-COT \cite{C-COT} &0.52 &51 &0.331 & 0.3\\
			~~pyMDNet \cite{MDNet} &- &- &0.304 &2\\
			~~SA-Siam \cite{SASiam} &0.53 &- &0.291 & 50\\
			~~StructSiam \cite{StructSiam} &- &- &0.264 &45\\
			~~MemTrack \cite{MemTrack} & 0.53 & - & 0.273 &50\\
			~~SiamFC \cite{SiamFc}  &0.53  &99 &0.235 &86\\	
			~~SCT \cite{SCT} &0.48 &117 &0.188 &40\\
			~~DSST \cite{DSST} &0.53 &151 &0.181&25\\	
			~~KCF \cite{KCF} &0.49 &122 &0.192 &170\\
			\hline
			\hline
			~~UDT (Ours) &0.54 &102 &0.226 &70\\
			~~UDT+ (Ours) &0.53 &66 &0.301 &55\\
			\hline
		\end{tabular*}
	\end{center}
	\vspace{-0.05in}
\end{table}

{\noindent \bf Temple-Color Dataset.} The Temple-Color \cite{TempleColor128} is a more challenging benchmark with 128 color videos. We compare our method with the state-of-the-art trackers illustrated in Sec.~\ref{state-of-the-art}. The propose UDT tracker performs favorably against SiamFC and CFNet as shown in Fig.~\ref{fig:8}.

{\noindent \bf VOT2016 Dataset.} Furthermore, we report the evaluation results on the VOT2016 benchmark \cite{VOT2016}. 
The expected average overlap (EAO) is the final metric for tracker ranking according to the VOT report \cite{VOTpami}.
As shown in Table~\ref{table:table2}, the performance of our UDT tracker is comparable with the baseline trackers (e.g., SiamFC). 
The improved UDT+ tracker performs favorably against state-of-the-art fully-supervised trackers including SA-Siam \cite{SASiam}, StructSiam \cite{StructSiam} and MemTrack \cite{MemTrack}.

\begin{figure}
	\centering
	\includegraphics[width=8.2cm]{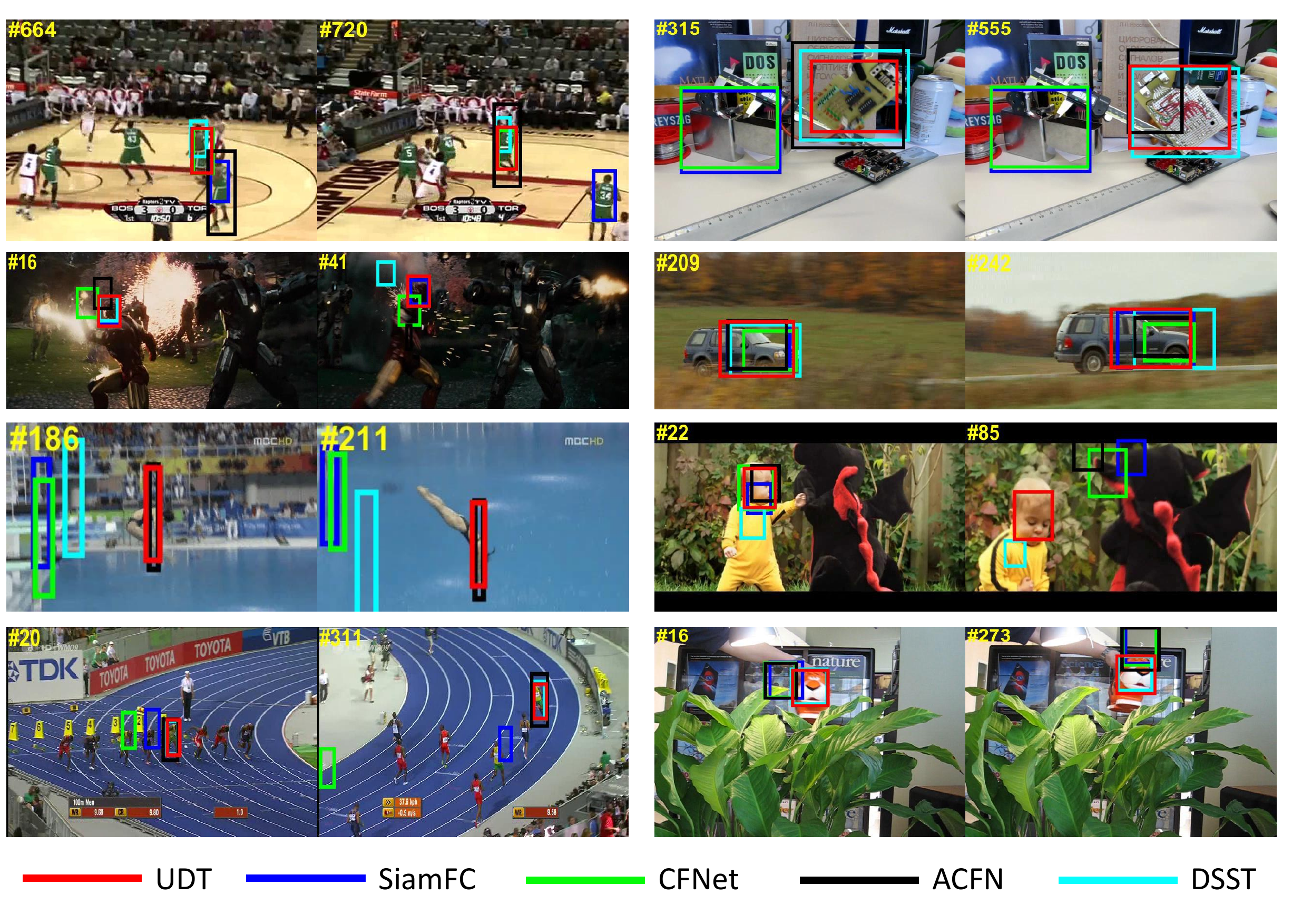}
	\caption{Qualitative evaluation of our proposed UDT and other trackers including SiamFC \cite{SiamFc}, CFNet \cite{CFNet}, ACFN \cite{ACFN}, and DSST \cite{DSST} on 8 challenging videos from OTB-2015. From left to right and top to down are \emph{Basketball}, \emph{Board}, \emph{Ironman}, \emph{CarScale}, \emph{Diving}, \emph{DragonBaby}, \emph{Bolt}, and \emph{Tiger1}, respectively.} \label{fig:10} \vspace{-0.02in}
\end{figure}

{\noindent \bf Attribute Analysis.} On the OTB-2015 benchmark, we further analyze the performance variations over different challenges as shown in Fig.~\ref{fig:9}. On the majority of challenging scenarios, the proposed UDT tracker outperforms the SiamFC and CFNet trackers. Compared with the fully-supervised UDT tracker, the unsupervised UDT does not achieve similar tracking accuracies under illumination variation (IV), occlusion (OCC), and fast motion (FM) scenarios. This is because the target appearance variations are significant in these video sequences. Without strong supervision, the proposed tracker is not effective to learn a robust feature representation to overcome these variations.

{\noindent \bf Qualitative Evaluation.} We visually compare the proposed UDT tracker to some supervised trackers (e.g., ACFN, SiamFC, and CFNet) and a baseline DCF tracker (DSST) on eight challenging video sequences. Although the proposed UDT tracker does not employ online improvements, we still observe that UDT effectively tracks the target, especially on the challenging \emph{Ironman} and \emph{Diving} video sequences as shown in Fig.~\ref{fig:10}. It is worth mentioning that such a robust tracker is learned using unlabeled videos without ground-truth supervisions.

{\noindent \bf Limitation.} (1) As discussed in the Attribute Analysis, our unsupervised feature representation may lack the objectness information to cope with complex scenarios. (2) Since our approach involves both forward and backward tracking, the computational load is another potential drawback.

\section{Conclusion}

In this paper, we proposed how to train a visual tracker using unlabeled video sequences in the wild, which has rarely been investigated in visual tracking.
By designing an unsupervised Siamese correlation filter network, we verified the feasibility and effectiveness of our forward-backward based unsupervised training pipeline.
To further facilitate the unsupervised training, we extended our framework to consider multiple frames and employ a cost-sensitive loss.
Extensive experiments exhibit that the proposed unsupervised tracker, without bells and whistles, performs as a solid baseline and achieves comparable results with the classic fully-supervised trackers.
%
%
Finally, unsupervised framework shows attractive potentials in visual tracking, such as utilizing more unlabeled data or weakly labeled data to further improve the tracking accuracy.

\footnotesize {\flushleft \bf Acknowledgements}. This work was supported in part to Dr. Houqiang Li by the 973 Program under Contract No. 2015CB351803 and NSFC under contract No. 61836011, and in part to Dr. Wengang Zhou by NSFC under contract No. 61822208 and 61632019, Young Elite Scientists Sponsorship Program By CAST (2016QNRC001), and the Fundamental Research Funds for the Central Universities. This work was supported in part by National Key Research and Development Program of China (2016YFB1001003), STCSM(18DZ1112300).

{\small
	\bibliographystyle{ieee}
	\bibliography{IEEEabrv,AgeGender}

\begin{thebibliography}{10}\itemsep=-1pt

\bibitem{SiamFc}
Luca Bertinetto, Jack Valmadre, Jo{\~a}o~F Henriques, Andrea Vedaldi, and
  Philip~HS Torr.
\newblock Fully-convolutional siamese networks for object tracking.
\newblock In {\em ECCV}, 2016.

\bibitem{MOSSE}
David~S Bolme, J~Ross Beveridge, Bruce~A Draper, and Yui~Man Lui.
\newblock Visual object tracking using adaptive correlation filters.
\newblock In {\em CVPR}, 2010.

\bibitem{VGGM}
Ken Chatfield, Karen Simonyan, Andrea Vedaldi, and Andrew Zisserman.
\newblock Return of the devil in the details: Delving deep into convolutional
  nets.
\newblock In {\em BMVC}, 2014.

\bibitem{ACT}
Boyu Chen, Dong Wang, Peixia Li, Shuang Wang, and Huchuan Lu.
\newblock Real-time'actor-critic'tracking.
\newblock In {\em ECCV}, 2018.

\bibitem{SCT}
Jongwon Choi, Hyung Jin~Chang, Jiyeoup Jeong, Yiannis Demiris, and Jin
  Young~Choi.
\newblock Visual tracking using attention-modulated disintegration and
  integration.
\newblock In {\em CVPR}, 2016.

\bibitem{ACFN}
Jongwon Choi, Hyung Jin~Chang, Sangdoo Yun, Tobias Fischer, Yiannis Demiris,
  and Jin Young~Choi.
\newblock Attentional correlation filter network for adaptive visual tracking.
\newblock In {\em CVPR}, 2017.

\bibitem{ECO}
Martin Danelljan, Goutam Bhat, Fahad Shahbaz~Khan, and Michael Felsberg.
\newblock Eco: Efficient convolution operators for tracking.
\newblock In {\em CVPR}, 2017.

\bibitem{DSST}
Martin Danelljan, Gustav H{\"a}ger, Fahad Khan, and Michael Felsberg.
\newblock Accurate scale estimation for robust visual tracking.
\newblock In {\em BMVC}, 2014.

\bibitem{SRDCFdecon}
Martin Danelljan, Gustav H{\"a}ger, Fahad~Shahbaz Khan, and Michael Felsberg.
\newblock Adaptive decontamination of the training set: A unified formulation
  for discriminative visual tracking.
\newblock In {\em CVPR}, 2016.

\bibitem{SRDCF}
Martin Danelljan, Gustav Hager, Fahad Shahbaz~Khan, and Michael Felsberg.
\newblock Learning spatially regularized correlation filters for visual
  tracking.
\newblock In {\em ICCV}, 2015.

\bibitem{C-COT}
Martin Danelljan, Andreas Robinson, Fahad~Shahbaz Khan, and Michael Felsberg.
\newblock Beyond correlation filters: Learning continuous convolution operators
  for visual tracking.
\newblock In {\em ECCV}, 2016.

\bibitem{Siamtriplet}
Xingping Dong and Jianbing Shen.
\newblock Triplet loss in siamese network for object tracking.
\newblock In {\em ECCV}, 2018.

\bibitem{HP}
Xingping Dong, Jianbing Shen, Wenguan Wang, Yu Liu, Ling Shao, and Fatih
  Porikli.
\newblock Hyperparameter optimization for tracking with continuous deep
  q-learning.
\newblock In {\em CVPR}, 2018.

\bibitem{Dsiam}
Qing Guo, Wei Feng, Ce Zhou, Rui Huang, Liang Wan, and Song Wang.
\newblock Learning dynamic siamese network for visual object tracking.
\newblock In {\em ICCV}, 2017.

\bibitem{SASiam}
Anfeng He, Chong Luo, Xinmei Tian, and Wenjun Zeng.
\newblock A twofold siamese network for real-time object tracking.
\newblock In {\em CVPR}, 2018.

\bibitem{KCF}
Jo{\~a}o~F Henriques, Rui Caseiro, Pedro Martins, and Jorge Batista.
\newblock High-speed tracking with kernelized correlation filters.
\newblock {\em TPAMI}, 37(3):583--596, 2015.

\bibitem{EAST}
Chen Huang, Simon Lucey, and Deva Ramanan.
\newblock Learning policies for adaptive tracking with deep feature cascades.
\newblock In {\em ICCV}, 2017.

\bibitem{Huangjianglei}
Jianglei Huang and Wengang Zhou.
\newblock Re2ema: Regularized and reinitialized exponential moving average for
  target model update in object tracking.
\newblock In {\em AAAI}, 2019.

\bibitem{RTMDNet}
Ilchae Jung, Jeany Son, Mooyeol Baek, and Bohyung Han.
\newblock Real-time mdnet.
\newblock In {\em ECCV}, 2018.

\bibitem{TLD}
Zdenek Kalal, Krystian Mikolajczyk, and Jiri Matas.
\newblock Tracking-learning-detection.
\newblock {\em TPAMI}, 34(7):1409--1422, 2012.

\bibitem{VOT2016}
Matej Kristan, Jiri Matas, Ales Leonardis, Michael Felsberg, Luka Cehovin,
  Gustavo Fern{\'a}ndez, Tomas Vojir, Hager, and et al.
\newblock The visual object tracking vot2016 challenge results.
\newblock In {\em ECCV Workshop}, 2016.

\bibitem{VOTpami}
Matej Kristan, Jiri Matas, Ale{\v{s}} Leonardis, Tom{\'a}{\v{s}}
  Voj{\'\i}{\v{r}}, Roman Pflugfelder, Gustavo Fernandez, Georg Nebehay, Fatih
  Porikli, and Luka {\v{C}}ehovin.
\newblock A novel performance evaluation methodology for single-target
  trackers.
\newblock {\em TPAMI}, 38(11):2137--2155, 2016.

\bibitem{Alexnet}
Alex Krizhevsky, Ilya Sutskever, and Geoffrey~E Hinton.
\newblock Imagenet classification with deep convolutional neural networks.
\newblock In {\em NIPS}, 2012.

\bibitem{le2013building}
Quoc~V Le.
\newblock Building high-level features using large scale unsupervised learning.
\newblock In {\em ICASSP}, 2013.

\bibitem{Multihypothesis}
Dae~Youn Lee, Jae~Young Sim, and Chang~Su Kim.
\newblock Multihypothesis trajectory analysis for robust visual tracking.
\newblock In {\em CVPR}, 2015.

\bibitem{lee2017unsupervised}
Hsin-Ying Lee, Jia-Bin Huang, Maneesh Singh, and Ming-Hsuan Yang.
\newblock Unsupervised representation learning by sorting sequences.
\newblock In {\em ICCV}, 2017.

\bibitem{SiamRPN}
Bo Li, Junjie Yan, Wei Wu, Zheng Zhu, and Xiaolin Hu.
\newblock High performance visual tracking with siamese region proposal
  network.
\newblock In {\em CVPR}, 2018.

\bibitem{IBCCF}
Feng Li, Yingjie Yao, Peihua Li, David Zhang, Wangmeng Zuo, and Ming-Hsuan
  Yang.
\newblock Integrating boundary and center correlation filters for visual
  tracking with aspect ratio variation.
\newblock In {\em ICCV Workshop}, 2017.

\bibitem{TempleColor128}
Pengpeng Liang, Erik Blasch, and Haibin Ling.
\newblock Encoding color information for visual tracking: algorithms and
  benchmark.
\newblock {\em TIP}, 24(12):5630--5644, 2015.

\bibitem{Liu_StructuralCF}
Si Liu, Tianzhu Zhang, Xiaochun Cao, and Changsheng Xu.
\newblock Structural correlation filter for robust visual tracking.
\newblock In {\em CVPR}, 2016.

\bibitem{DSLT}
Xiankai Lu, Chao Ma, Bingbing Ni, Xiaokang Yang, Ian Reid, and Ming-Hsuan Yang.
\newblock Deep regression tracking with shrinkage loss.
\newblock In {\em ECCV}, 2018.

\bibitem{CSR-DCF}
Alan Lukezic, Tomas Vojir, Luka Cehovin~Zajc, Jiri Matas, and Matej Kristan.
\newblock Discriminative correlation filter with channel and spatial
  reliability.
\newblock In {\em CVPR}, 2017.

\bibitem{luo2019end}
Wenhan Luo, Peng Sun, Fangwei Zhong, Wei Liu, Tong Zhang, and Yizhou Wang.
\newblock End-to-end active object tracking and its real-world deployment via
  reinforcement learning.
\newblock {\em TPAMI}, 2019.

\bibitem{luo2014multiple}
Wenhan Luo, Junliang Xing, Anton Milan, Xiaoqin Zhang, Wei Liu, Xiaowei Zhao,
  and Tae-Kyun Kim.
\newblock Multiple object tracking: A literature review.
\newblock {\em arXiv preprint arXiv:1409.7618}, 2014.

\bibitem{HCF}
Chao Ma, Jia-Bin Huang, Xiaokang Yang, and Ming-Hsuan Yang.
\newblock Hierarchical convolutional features for visual tracking.
\newblock In {\em ICCV}, 2015.

\bibitem{2018trackingnet}
Matthias M{\"u}ller, Adel Bibi, Silvio Giancola, Salman Al-Subaihi, and Bernard
  Ghanem.
\newblock Trackingnet: A large-scale dataset and benchmark for object tracking
  in the wild.
\newblock In {\em ECCV}, 2018.

\bibitem{MDNet}
Hyeonseob Nam and Bohyung Han.
\newblock Learning multi-domain convolutional neural networks for visual
  tracking.
\newblock In {\em CVPR}, 2016.

\bibitem{olshausen1997sparse}
Bruno~A Olshausen and David~J Field.
\newblock Sparse coding with an overcomplete basis set: A strategy employed by
  v1?
\newblock {\em Vision research}, 37(23):3311--3325, 1997.

\bibitem{DATnips}
Shi Pu, Yibing Song, Chao Ma, Honggang Zhang, and Ming-Hsuan Yang.
\newblock Deep attentive tracking via reciprocative learning.
\newblock In {\em NeurIPS}, 2018.

\bibitem{HDT}
Yuankai Qi, Shengping Zhang, Lei Qin, Hongxun Yao, Qingming Huang, and Jongwoo
  Lim Ming-Hsuan Yang.
\newblock Hedged deep tracking.
\newblock In {\em CVPR}, 2016.

\bibitem{youtubeBB}
Esteban Real, Jonathon Shlens, Stefano Mazzocchi, Xin Pan, and Vincent
  Vanhoucke.
\newblock Youtube-boundingboxes: A large high-precision human-annotated data
  set for object detection in video.
\newblock In {\em CVPR}, 2017.

\bibitem{ILSVRC2015}
Olga Russakovsky, Jia Deng, Hao Su, Jonathan Krause, Sanjeev Satheesh, Sean Ma,
  Zhiheng Huang, Andrej Karpathy, Aditya Khosla, Michael Bernstein, et~al.
\newblock Imagenet large scale visual recognition challenge.
\newblock {\em IJCV}, 115(3):211--252, 2015.

\bibitem{VGG}
Karen Simonyan and Andrew Zisserman.
\newblock Very deep convolutional networks for large-scale image recognition.
\newblock {\em arXiv preprint arXiv:1409.1556}, 2014.

\bibitem{CREST}
Yibing Song, Chao Ma, Lijun Gong, Jiawei Zhang, Rynson Lau, and Ming-Hsuan
  Yang.
\newblock Crest: Convolutional residual learning for visual tracking.
\newblock In {\em ICCV}, 2017.

\bibitem{VITAL}
Yibing Song, Chao Ma, Xiaohe Wu, Lijun Gong, Linchao Bao, Wangmeng Zuo, Chunhua
  Shen, Rynson~W.H. Lau, and Ming-Hsuan Yang.
\newblock Vital: Visual tracking via adversarial learning.
\newblock In {\em CVPR}, 2018.

\bibitem{SINT}
Ran Tao, Efstratios Gavves, and Arnold~WM Smeulders.
\newblock Siamese instance search for tracking.
\newblock In {\em CVPR}, 2016.

\bibitem{KLTtracker}
Carlo Tomasi and Takeo Kanade.
\newblock Detection and tracking of point features.
\newblock 1991.

\bibitem{2018longtermBenchmark}
Jack Valmadre, Luca Bertinetto, Jo{\~a}o~F Henriques, Ran Tao, Andrea Vedaldi,
  Arnold Smeulders, Philip Torr, and Efstratios Gavves.
\newblock Long-term tracking in the wild: A benchmark.
\newblock In {\em ECCV}, 2018.

\bibitem{CFNet}
Jack Valmadre, Luca Bertinetto, Jo{\~a}o~F Henriques, Andrea Vedaldi, and
  Philip~HS Torr.
\newblock End-to-end representation learning for correlation filter based
  tracking.
\newblock In {\em CVPR}, 2017.

\bibitem{vondrick2016anticipating}
Carl Vondrick, Hamed Pirsiavash, and Antonio Torralba.
\newblock Anticipating visual representations from unlabeled video.
\newblock In {\em CVPR}, 2016.

\bibitem{DLT}
Naiyan Wang and Dit-Yan Yeung.
\newblock Learning a deep compact image representation for visual tracking.
\newblock In {\em NIPS}, 2013.

\bibitem{ningwangTCSVT}
Ning Wang, Wengang Zhou, and Houqiang Li.
\newblock Reliable re-detection for long-term tracking.
\newblock {\em TCSVT}, 2019.

\bibitem{MCCT}
Ning Wang, Wengang Zhou, Qi Tian, Richang Hong, Meng Wang, and Houqiang Li.
\newblock Multi-cue correlation filters for robust visual tracking.
\newblock In {\em CVPR}, 2018.

\bibitem{DCFNet}
Qiang Wang, Jin Gao, Junliang Xing, Mengdan Zhang, and Weiming Hu.
\newblock Dcfnet: Discriminant correlation filters network for visual tracking.
\newblock {\em arXiv preprint arXiv:1704.04057}, 2017.

\bibitem{RASNet}
Qiang Wang, Zhu Teng, Junliang Xing, Jin Gao, Weiming Hu, and Stephen Maybank.
\newblock Learning attentions: Residual attentional siamese network for high
  performance online visual tracking.
\newblock In {\em CVPR}, 2018.

\bibitem{wang2015unsupervised}
Xiaolong Wang and Abhinav Gupta.
\newblock Unsupervised learning of visual representations using videos.
\newblock In {\em ICCV}, 2015.

\bibitem{OTB-2015}
Yi Wu, Jongwoo Lim, and Ming-Hsuan Yang.
\newblock Object tracking benchmark.
\newblock {\em TPAMI}, 37(9):1834--1848, 2015.

\bibitem{MemTrack}
Tianyu Yang and Antoni~B Chan.
\newblock Learning dynamic memory networks for object tracking.
\newblock In {\em ECCV}, 2018.

\bibitem{SACF}
Mengdan Zhang, Qiang Wang, Junliang Xing, Jin Gao, Peixi Peng, Weiming Hu, and
  Steve Maybank.
\newblock Visual tracking via spatially aligned correlation filters network.
\newblock In {\em ECCV}, 2018.

\bibitem{StructSiam}
Yunhua Zhang, Lijun Wang, Jinqing Qi, Dong Wang, Mengyang Feng, and Huchuan Lu.
\newblock Structured siamese network for real-time visual tracking.
\newblock In {\em ECCV}, 2018.

\end{thebibliography}
}

\end{document}